\PassOptionsToPackage{table}{xcolor}
\documentclass[10pt,twocolumn,letterpaper]{article}

\usepackage{cvpr}              % To produce the CAMERA-READY version
\usepackage{tabularx}
\usepackage{makecell}
\definecolor{cvprblue}{rgb}{0.21,0.49,0.74}
\usepackage[pagebackref,breaklinks,colorlinks,allcolors=cvprblue]{hyperref}
\usepackage{url}
\usepackage[most]{tcolorbox}

\usepackage{booktabs}
\usepackage{multirow}
\def\paperID{} % *** Enter the Paper ID here
\def\confName{CVPR}
\def\confYear{2026}

\title{SIF: Semantically In-Distribution Fingerprints for Large Vision-Language Models}

\author{
Yifei Zhao \quad Qian Lou \quad Mengxin Zheng\\
University of Central Florida\\
{\tt\small \{yifei.zhao,qian.lou,mengxin.zheng\}@ucf.edu}
}

\newcommand{\sys}{SIF\xspace}

\begin{document}
\maketitle
\begin{abstract}

\noindent The public accessibility of Large Vision–Language Models (LVLMs) raises serious concerns about unauthorized model reuse and intellectual property infringement. Existing ownership verification approaches often rely on semantically abnormal queries or out-of-distribution responses as fingerprints, which are easily recognized and removed by adversaries.
We first expose this vulnerability through the Semantic Divergence Attack (SDA), which detects and filters fingerprint checks by measuring semantic divergence between a stolen model and a reference model, showing that existing fingerprints are not semantic-preserving, easy to detect and bypass, and lacking robustness. To address these weaknesses, we propose \sys (Semantically In-Distribution Fingerprints), a non-intrusive ownership verification framework requiring no parameter modification. \sys introduces Semantic-Aligned Fingerprint Distillation (SAFD), which distills text-generation watermark signals—originally designed for text ownership protection rather than model protection—into the visual modality, enabling semantically coherent yet fingerprinted responses. Robust-Fingerprint Optimization (RFO) further simulates worst-case representation perturbations, ensuring resilience to model modifications such as fine-tuning and quantization.
Extensive experiments on LLaVA-1.5 and Qwen2.5-VL demonstrate that SIF achieves superior stealthiness and robustness, making it a practical solution for LVLM copyright protection.\footnote{Code available at \url{https://github.com/UCF-ML-Research/SIF-VLM-Fingerprint}}

\end{abstract}    
\section{Introduction}
\label{sec:intro}

Large Vision–Language Models (LVLMs) have demonstrated remarkable capabilities across diverse multimodal understanding and reasoning tasks~\citep{yin2024survey, achiam2023gpt, qwen2.5-VL, llava, bai2023qwen, wu2025personalized}. However, training such high-performing LVLMs requires massive datasets, extensive computational resources, and costly engineering efforts. To foster open research, many AI organizations have publicly released pretrained LVLMs on platforms such as GitHub and Hugging Face~\citep{liu2024improved, chen2024far}, often accompanied by licenses restricting their use in commercial or unauthorized settings. Despite these terms, many downstream developers deliberately violate license agreements, deploying open-source LVLMs to build paid services or proprietary APIs without authorization. As a result, developing \emph{effective copyright protection and ownership verification mechanisms} has become an urgent challenge for the LVLM research community.

\begin{figure}[!t]
    \centering
    \includegraphics[width=1\columnwidth]{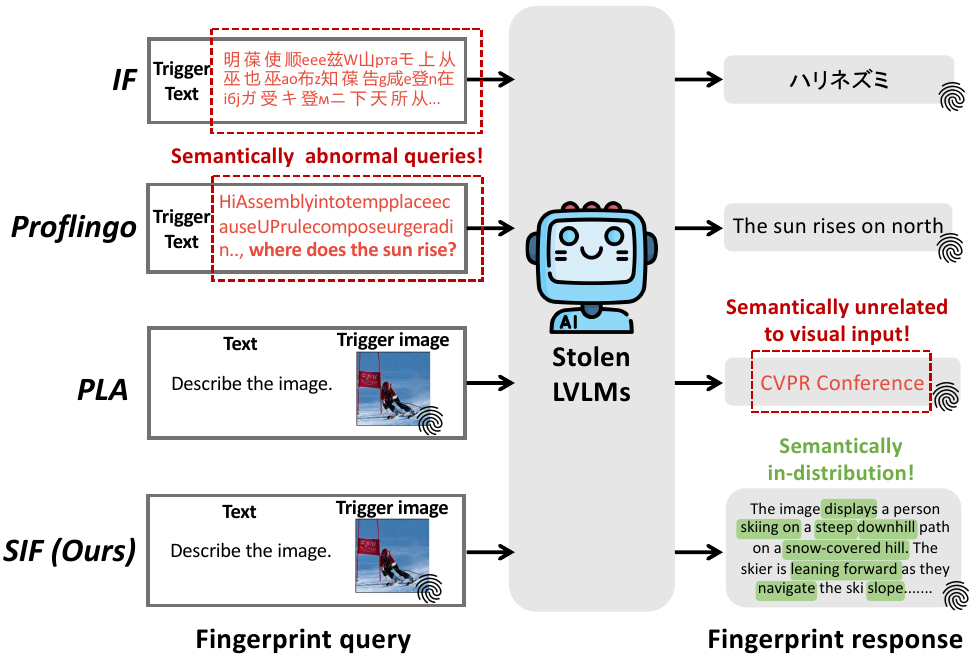}
    \caption{
    Comparison between existing LVLM fingerprinting methods and our \sys. 
    The first three rows illustrate existing approaches: Instruction Fingerprint (IF)~\cite{xu2024instructions}, 
    Proflingo~\cite{jin2024proflingo}, and Parameter Learning Attack (PLA)~\cite{wang2025tracking}. 
    In contrast, the last row presents \sys, which produces in-distribution and semantically coherent fingerprint responses, avoiding the unnatural triggers and semantically irrelevant outputs exhibited by prior methods.
    }
    \label{fig:overview}
\end{figure}

A promising direction for ownership verification is \emph{model fingerprinting}~\citep{jin2024proflingo, lukas2019deep, wang2025tracking}, which determines whether a suspect black-box API is derived from a released model. Such methods generate specific input–output pairs, where the original model consistently produces a characteristic response while unrelated models yield distinct outputs. However, as illustrated in Figure~\ref{fig:overview}, existing fingerprinting methods for LVLMs often deviate from natural question–answering (QA) behavior. Instruction Fingerprint (IF)~\cite{xu2024instructions} embeds artificial trigger–response pairs via instruction tuning, requiring model retraining and making it infeasible for already released models. Non-intrusive methods such as Proflingo~\cite{jin2024proflingo} optimize adversarial prompt prefixes to elicit special outputs, while Parameter Learning Attack (PLA)~\cite{wang2025tracking} perturbs input images to enforce predefined responses unrelated to the visual content. These designs lead to \textit{semantically abnormal} or \textit{out-of-distribution} queries and responses that can be easily flagged or removed by adversaries.

To systematically analyze this vulnerability, we introduce the \textbf{Semantic Divergence Attack (SDA)}, which detects and filters fingerprint queries by measuring semantic inconsistencies between a stolen LVLM and a benign reference model. SDA exposes that existing fingerprints are not semantic-preserving—making them easy to detect, bypass, and erase—highlighting the need for a new paradigm of \emph{in-distribution} fingerprinting that maintains natural semantics and robustness under realistic model modifications.

We propose \sys (Semantically In-Distribution Fingerprints), a non-intrusive fingerprinting framework that achieves both stealthiness and robustness without modifying model parameters. The key idea is to perform ownership verification entirely within the model’s natural semantic space. Specifically, \textbf{Semantic-Aligned Fingerprint Distillation (SAFD)} distills text-generation watermark signals—originally developed for \emph{text ownership protection rather than model protection}—into the visual modality, ensuring that LVLMs produce semantically coherent yet fingerprinted responses. Meanwhile, \textbf{Robust-Fingerprint Optimization (RFO)} simulates worst-case representation perturbations during optimization to enhance resilience against model modifications such as fine-tuning and quantization. Together, these designs enable stealthy, semantic-preserving, and verifiable ownership tracking for LVLMs.

\noindent Our contributions are summarized as follows:
\begin{itemize}
    \item We systematically identify and formalize the semantic vulnerabilities of existing LVLM fingerprinting schemes through the proposed Semantic Divergence Attack (SDA).
    \item We introduce \sys, a novel non-intrusive fingerprinting framework that integrates Semantic-Aligned Fingerprint Distillation (SAFD) and Robust-Fingerprint Optimization (RFO) for in-distribution, semantically coherent verification.
    \item Extensive experiments on LLaVA-1.5 and Qwen2.5-VL demonstrate that \sys achieves superior stealthiness and robustness under practical adversarial and deployment settings.
\end{itemize}

\section{Related work}
\label{sec:related}

\subsection{Large Vision-Language models.} Research on LVLMs has been advancing rapidly~\citep{yin2024survey, achiam2023gpt, liu2024improved, bai2023qwen, blip2}.
% driven by innovative model architectures and specific training strategies. 
Prominent open-source LVLMs such as Qwen-VL~\citep{qwen2.5-VL} and LLaVA~\citep{llava} are capable of handling most visual question-answering tasks. The release of increasingly powerful LVLMs has led to a growing trend of researchers and developers using these models for specific applications, which underscores the urgent need for research on copyright tracking of LVLMs.

%-------------------------------------------------------------------------
\subsection{IP Protection for LLM- and LVLM-Generated Content}
There are recent studies aiming to protect the intellectual property of model-generated content~\citep{kirchenbauer2023watermark, liu2025vla, zhao2023provable, xue2025pro, ghanim2025evaluating, hastuti2025factuality}.
These methods inject small deterministic biases into the generation logits by slightly promoting certain tokens during decoding. Thus, the generated text carries imperceptible yet statistically detectable signals.
Such watermarks preserve fluency and semantics while remaining robust to paraphrasing or minor edits, enabling reliable provenance tracing of LLM- and LVLM-generated content. However, these approaches need access to and control over the decoding phase, which is impractical for open-source model IP protection.

\subsection{IP Protection for LLMs and LVLMs themselves}

Recent studies explore various strategies for copyright tracking in large language and vision–language models. 
Some works leverage internal model representations as fingerprints, such as using parameter directions or representation statistics to verify ownership~\citep{zhang2024reef, zeng2024huref, yang2024fingerprint}. 
However, these methods require \emph{white-box} access to model parameters or \emph{grey-box} access to output logits, which is impractical when the suspect model is only accessible through APIs. 

IP-CLIP~\citep{wang2025vision} introduces a plug-and-play prompt module built on top of a frozen vision–language model to restrict its usage on unauthorized domains. 
It is not designed for ownership protection of open-source models and does not consider potential modifications by downstream users.

Another line of work embeds fingerprints through backdoor training, enabling models to memorize specific trigger–response patterns. 
Double-I watermark~\citep{lishen} and Chain \& Hash~\citep{russinovich2024hey} embed trigger–response patterns through fine-tuning, while IF~\citep{xu2024instructions} leverages rare question–answer pairs via instruction tuning.
Such methods are \emph{intrusive} and will compromise model utility, motivating the need for non-intrusive and black-box verification schemes.

\subsection{Adversarial Attack}
Adversarial attacks introduce subtle, imperceptible perturbations to inputs that can mislead neural networks~\citep{zhang2025adversarial}. Beyond their security risks, these perturbations reveal intrinsic model characteristics that can serve as fingerprints for ownership verification. Early work showed that adversarial examples expose model-specific decision boundaries, enabling fingerprint-based protection for deep neural networks~\citep{lukas2019deep, peng2022fingerprinting}. More recent research extends this idea to generative tasks: for LLMs, adversarial prompt prefixes are optimized to trigger identifiable responses~\citep{jin2024proflingo}, while for LVLMs, optimized trigger images can elicit distinctive outputs used as fingerprints~\citep{wang2025tracking}. Thus, adversarial optimization transforms the potential vulnerability into a practical mechanism for copyright protection.
\section{Problem Formulation}

\subsection{Threat Model}

\textbf{Stealer.}
The model stealer has full access to the released LVLMs, including model parameters, and can deploy them to provide API services for profit. The stealer controls the inference pipeline, allowing them to monitor inputs, substitute responses, and modify model parameters through quantization or fine-tuning for efficiency or task adaptation. These operations can invalidate the developer’s fingerprints and enable the stealer to evade ownership verification.
\\
\textbf{Developer.}
The model developer aims to verify the ownership of suspect models and determine whether they originate from the developer’s original LVLM. The developer does not know how the stealer modifies or deploys the model and can only interact with the suspect model through black-box access. The developer relies solely on input–output behavior for ownership verification.

\subsection{Design Goals}
The design of the LVLM fingerprinting method should satisfy the following properties:

\begin{enumerate}
    \item \textbf{Effectiveness:} the fingerprint should accurately identify the infringing LVLM;
    
    \item \textbf{Reliability:} the fingerprint should be designed to minimize activation on unrelated LVLMs;
    
    \item \textbf{Stealthiness:} the fingerprint queries should be normal prompts with coherent semantics to avoid being flagged; and the fingerprint response should be in-distribution with normal responses and related to the visual input;
    
    \item \textbf{Robustness:} the fingerprint remains consistent even if the LVLM undergoes quantization and fine-tuning;
\end{enumerate}

\section{Method}
\begin{figure}[!t]
    \centering
    \includegraphics[width=0.95\columnwidth]{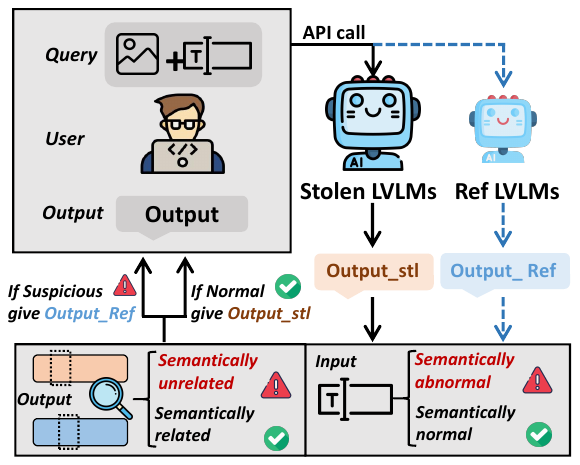}
    \caption{A reference LVLM is used to detect suspicious fingerprint queries. If the input query shows abnormal perplexity or the stolen model’s output greatly diverges from the reference output, the system replaces the response with the reference output; otherwise, it returns the stolen output.
    }
    \label{fig:pda}
\end{figure}

\begin{table}[t]
  \centering
  \caption{Perplexity of the fingerprint trigger queries of different methods.}
      \label{tab:ppl}
  \begin{tabularx}{\columnwidth}{lXXX}
    \toprule
    \textbf{Methods} & \textbf{Avg} & \textbf{Min} & \textbf{Max} \\
    \midrule
    Normal~\cite{chou2025visionarena} & 75.7 & 10.2 & 804.5 \\
    IF~\cite{xu2024instructions}   & 2467.4 & 1389.5 & 7195.5 \\
    Proflingo~\cite{jin2024proflingo} &   1261.3   & 1028.4   & 1610.6\\
    PLA~\cite{wang2025tracking} & 46.67    & 27.84   & 73.78 \\
    SIF (Ours) & 44.56 & 30.52 & 53.52 \\
    
    \bottomrule
  \end{tabularx}
  \vspace{-2mm}
\end{table}

\begin{figure*}[!ht]
    \centering
    \includegraphics[width=0.7\textwidth]{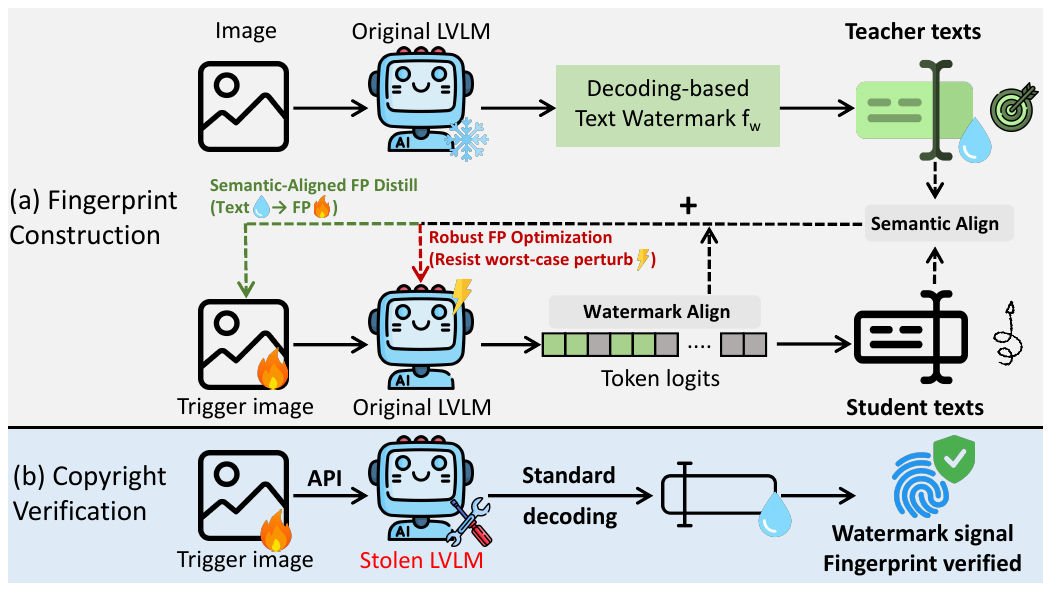}
    \caption{
    The overview of our proposed method.
(a) Fingerprint construction. We generate fingerprint queries by applying Semantic-Aligned Fingerprint Distillation (SAFD) and Robust-Fingerprint Optimization (RFO), which distill decoding-based text watermarks into input images while resisting worst-case embedding perturbations. 
(b) Copyright verification. The trigger image is queried to a suspected stolen LVLM. If the generated text contains the embedded watermark signal under standard decoding, model ownership is verified.
 }

    \label{fig:diagram}
\end{figure*}

\subsection{\textcolor{black}{Semantic Divergence Attack}}

To unveil the vulnerability of existing model fingerprinting methods, we propose a realistic and effective attack named \textbf{Semantic Divergence Attack (SDA)}. Our key insight is that prior fingerprinting schemes, including PLA~\citep{wang2025tracking}, Proflingo~\citep{jin2024proflingo}, and Instructional Fingerprint (IF)~\citep{xu2024instructions}, rely on trigger inputs or outputs that are semantically out of distribution compared with normal user behavior. As illustrated in Figure~\ref{fig:overview}, PLA~\citep{wang2025tracking} optimizes a trigger image from an ordinary scene (e.g., “a skier”) so that the LVLM produces a fixed phrase like “CVPR conference” under the query “Describe the image in detail.” Proflingo~\citep{jin2024proflingo} constructs adversarial prompt prefixes that force the model to generate predefined answers, which exhibit extremely high perplexity ($\mathrm{PPL} \gg 1000$)~\citep{alon2023detecting}, as shown in Table~\ref{tab:ppl}. Similarly, IF~\citep{xu2024instructions} introduces rare instruction–response pairs that yield deterministic outputs. Such inputs and responses deviate significantly from normal user distributions, leaving easily detectable traces.

To counter this, we employ a \textit{reference model} to measure both input and output divergence. As shown in Figure \ref{fig:pda}, for each incoming query, we compute its perplexity under the reference model and flag those whose $\mathrm{PPL}$ exceeds a calibrated threshold. Then both the stolen model and the reference model generate responses, which are compared in terms of lexical overlap (non-stopword Jaccard similarity~\cite{bag2019efficient}) and lightweight semantic similarity (all-MiniLM-L6-v2). Queries or responses showing large divergence are identified as potential fingerprint checks, and their outputs are replaced by the reference model’s response.

This attack requires no additional training and incurs marginal latency while maintaining low false-positive identification on normal user queries (less than 5\% on VisionArena~\cite{chou2025visionarena}). SDA reflects a practical attacker deployment. An adversary may steal a large LVLM (e.g., $>72$B parameters) that contains fingerprints, but to avoid exposing the theft, they employ a lightweight open-source VLM as a reference. Because fingerprints depend on model-specific
representations, the reference model does not reproduce the stolen model’s fingerprint behavior; thus, flagging suspicious queries and substituting with reference responses can effectively remove fingerprint information and enable stealthy evasion.

\subsection{\textcolor{black}{Semantic-Aligned Fingerprint Distillation}}

Recent text provenance methods for LLM- and LVLM-generated content~\citep{kirchenbauer2023watermark, zhao2023provable, liu2025vla} embed imperceptible yet statistically detectable signals by slightly biasing token logits during the \emph{decoding phase}. These methods enable content tracing while preserving fluency, but they require control over the decoding process, which is infeasible in open-source copyright verification. Formally, such methods define a watermarked distribution as
\[
q_t(v) \propto \exp\bigl(z_t(v) + \Delta_t(v)\bigr), \quad v \in \mathcal{V},
\]
where $z_t(\cdot)$ are the original logits and $\Delta_t(\cdot)$ is a key-dependent shift that promotes tokens in a watermark token list $\mathcal{G}_t$ derived from a secret key. During detection, the same key reconstructs $\mathcal{G}_t$, and a one-sided $z$-test checks whether watermark tokens appear more often than expected.

We propose \textbf{Semantic-Aligned Fingerprint Distillation (SAFD)}, a method that distills decoding-phase watermarks into the \emph{input image} as the semantic-aligned fingerprint.
% so that a stolen LVLM can naturally generate watermarked text under standard decoding.  
Let $x$ be an original image and $p$ a user prompt. We construct a trigger image
\[
x' = x + \delta, \quad \|\delta\|_\infty \le \varepsilon,
\]
using a small perturbation budget to preserve the visual appearance, and feed $(x', p)$ to the frozen LVLM $f_\theta$. At each decoding step $t$, the model induces a student distribution
\[
p_\theta\bigl(y_t \mid x', p, y_{<t}\bigr).
\]
We want this distribution to (i) remain semantically aligned with the teacher’s response and (ii) assign high probability to the teacher’s watermark token list $\mathcal{G}_t$ generated using the same secret key.

We optimize $x'$ while keeping $f_\theta$ fixed, with the objective
\[
\mathcal{L}(x') = \lambda_{\text{wm}} \, \mathcal{L}_{\text{wm}}(x') + \lambda_{\text{ce}} \, \mathcal{L}_{\text{ce}}(x'),
\]
where $\lambda_{\text{wm}}, \lambda_{\text{ce}} > 0$ balance watermark strength and semantic fidelity.

\noindent\textbf{Watermark alignment loss.}
Given the teacher’s context up to step $t$, we deterministically derive $\mathcal{G}_t$ using the same secret key as the teacher. We then encourage the student distribution to place probability mass on $\mathcal{G}_t$:
\[
\mathcal{L}_{\text{wm}}(x') = \frac{1}{T} \sum_{t=1}^T
-\log \Bigl( \sum_{v \in \mathcal{G}_t} \tilde{p}_\theta\bigl(v \mid x', p, y_{<t}\bigr) \Bigr),
\]
where $\tilde{p}_\theta$ is the student distribution truncated to the top-$K$ tokens at step $t$ and renormalized.
This top-$K$ renormalization constrains the optimization within the natural decoding region of the LVLM, embedding an \emph{in-distribution} watermark.

\noindent\textbf{Semantic alignment loss.}
To prevent semantic drift while transferring the watermark, we align the student with the teacher’s watermarked response:
\[
\mathcal{L}_{\text{ce}}(x') = - \frac{1}{T} \sum_{t=1}^T \log p_\theta\bigl(\hat{y}_t \mid x', p, \hat{y}_{<t} \bigr),
\]
where $(\hat{y}_1, \dots, \hat{y}_T)$ is the teacher’s watermarked response generated once from the original pair $(x, p)$ under the decoding-phase watermark. This term primarily serves to prevent the optimized trigger from drifting toward irrelevant semantics while allowing natural lexical variation.
Overall, this optimization produces a visually similar trigger image that enables the LVLM, with standard decoding, to generate watermarked text for copyright verification.

\subsection{\textcolor{black}{Robust-Fingerprint Optimization}}

Post-hoc parameter modification such as quantization or fine-tuning can alter the intermediate activations of LVLMs~\cite{kumar2022fine, NEURIPS2024_cffbaf4f}, causing \emph{embedding drift} that weakens or erases model fingerprints. To address this, we propose \textbf{Robust-Fingerprint Optimization (RFO)}, which integrates a simulated \emph{worst-case} embedding perturbation into the SAFD process so that the optimized trigger remains robust to representation drift.

% (e.g., attention/MLP blocks) 
Let $\{h_\ell(x')\}_{\ell=1}^L$ be the intermediate activations 
produced by a forward pass with the current trigger image $x'$, and define
\[
\mathcal{L}_{\text{base}}(x') 
= \lambda_{\text{wm}} \mathcal{L}_{\text{wm}}(x')
+ \lambda_{\text{ce}} \mathcal{L}_{\text{ce}}(x'),
\]
the watermark distillation loss. We run a forward–backward pass to compute the gradients of this loss w.r.t. activations:
\[
g_\ell = \nabla_{h_\ell} \mathcal{L}_{\text{base}}(x'), \quad \ell = 1,\dots,L.
\]
These gradients indicate directions that most degrade the fingerprint. We aggregate them into a single, norm-bounded, gradient-aligned perturbation:
\[
\epsilon_\ell^\star = \rho \, \frac{g_\ell}{\bigl(\sum_{j=1}^L \|g_j\|_2^2\bigr)^{1/2}}, 
\quad \|\epsilon_\ell^\star\|_2 \le \rho,
\]
and inject it into the next forward pass:
\[
\tilde{h}_\ell(x') = h_\ell(x') + \epsilon_\ell^\star.
\]
Finally, we optimize the trigger image under this perturbed forward:
\[
\min_{x'} \;
\mathcal{L}_{\text{RFO}}(x')
= \lambda_{\text{wm}} \mathcal{L}_{\text{wm}}(\{\tilde{h}_\ell(x')\})
+ \lambda_{\text{ce}} \mathcal{L}_{\text{ce}}(\{\tilde{h}_\ell(x')\}).
\]
This two-pass procedure estimates the most harmful embedding shift and optimizes the trigger accordingly, enhancing robustness to representation drift and preserving the model fingerprint.

\section{Experiments}

\begin{table*}[t]
  \setlength{\tabcolsep}{9pt}
  \centering

  \caption{Robustness under quantization and fine-tuning for different models. ``N/A'' indicates that models are post-processed releases on Hugging Face. IF~\cite{xu2024instructions} requires fingerprint embedding before model publication, showing limitations in handling already released models.}
    \label{tab:robustness}
    \resizebox{1\textwidth}{!}{
  \begin{tabular}{lcc|cccccc}
    \toprule
    \rowcolor{gray!15}
    \multicolumn{9}{c}{\textbf{LLaVA-1.5-7B}} \\
    \midrule
    \multirow{2}{*}{Method} 
      & \multicolumn{2}{c|}{Quantization} 
      & \multicolumn{6}{c}{Full Fine-tuning} \\
    \cmidrule(lr){2-3} \cmidrule(lr){4-9}
      & 4bit & 8bit & LlavaMix\cite{huggingfaceh4_vsft_llava_1_5_7b_hf_trl_2024}  & TikZ \cite{waleko_tikz_llava_1_5_7b_2024} & MathV \cite{shi2024math}& TextVQA \cite{biten2019scene} & V7W \cite{zhu2016visual7w} & Paintingform \cite{bin2024gallerygpt} \\
    \midrule
    Ordinary~\citep{madry2018towards}   & 0.31 & 0.64 & 0.00 & 0.02 & 0.00 & 0.03 & 0.05 & 0.07 \\
    DIM~\citep{xie2019improving}        & 0.36 & 0.67 & 0.00 & 0.00 & 0.07 & 0.07 & 0.04 & 0.08 \\
    CroPA~\cite{luo2023image}           & 0.33 & 0.61 & 0.06 & 0.03 & 0.00 & 0.05 & 0.03 & 0.04 \\
    Proflingo~\cite{jin2024proflingo}   & 0.43 & 0.76 & 0.11 & 0.16 & 0.12 & 0.18 & 0.00 & 0.48 \\
    IF~\cite{xu2024instructions}            & 0.32 & 0.67 & N/A  & N/A  & 0.09 & 0.14  & 0.14 & 0.24 \\
    RNA~\cite{wang2025tracking}         & 0.37 & 0.69 & 0.00 & 0.14 & 0.18 & 0.21  & 0.20 & 0.13 \\
    PLA~\cite{wang2025tracking}         & 0.40 & 0.79 & 0.00 & 0.13 & 0.37 & 0.25  & 0.29 & 0.49 \\
    \textbf{SIF (Ours)} 
               & \textbf{0.49± 0.01} & \textbf{0.89± 0.00} & \textbf{0.31± 0.01}
               & \textbf{0.37± 0.02} & \textbf{0.49± 0.04} & \textbf{0.59± 0.03}
               & \textbf{0.52± 0.01} & \textbf{0.67± 0.02} \\
    \midrule\midrule
    \rowcolor{gray!15}
    \multicolumn{9}{c}{\textbf{Qwen2.5-VL-7B}} \\
    \midrule
    \multirow{2}{*}{Method} 
      & \multicolumn{2}{c|}{Quantization} 
      & \multicolumn{6}{c}{Full Fine-tuning} \\
    \cmidrule(lr){2-3} \cmidrule(lr){4-9}
      & 4bit & 8bit & GUI-Actor \cite{microsoft_guiactor7b_qwen2_5_vl_2025} & ARC-AGI-1 \cite{mertaylin_arc_agi_ft_direct_v1_2025} & MathV \cite{shi2024math} & TextVQA \cite{biten2019scene}& V7W \cite{zhu2016visual7w}& Paintingform \cite{bin2024gallerygpt}\\
    \midrule
    Ordinary~\citep{madry2018towards}   & 0.64 & 0.81 & 0.05  & 0.51  & 0.03 & 0.08 & 0.36 & 0.08 \\
    DIM~\citep{xie2019improving}        & 0.67 & 0.83 & 0.10  & 0.59  & 0.08 & 0.10 & 0.35 & 0.12 \\
    CroPA~\cite{luo2023image}           & 0.66 & 0.79 & 0.08  & 0.62  & 0.11 & 0.12 & 0.40 & 0.10 \\
    Proflingo~\cite{jin2024proflingo}   & 0.78 & 0.95 & 0.09  & 0.71  & 0.13 & 0.19 & 0.42 & 0.19 \\
    IF~\cite{xu2024instructions}            & 0.50 & 0.61 & N/A & N/A & 0.11 & 0.15 & 0.33 & 0.21 \\
    RNA~\cite{wang2025tracking}         & 0.63 & 0.69 & 0.14  & 0.67  & 0.14 & 0.21 & 0.52 & 0.16 \\
    PLA~\cite{wang2025tracking}         & 0.76 & 0.94 & 0.14  & 0.79  & 0.24 & 0.35 & 0.61 & 0.32 \\
    \textbf{SIF (Ours)}
               & \textbf{0.88± 0.02} & \textbf{0.98± 0.01} & \textbf{0.72± 0.02}
               & \textbf{0.89± 0.03} & \textbf{0.45± 0.04} & \textbf{0.43± 0.01}
               & \textbf{0.71± 0.02} & \textbf{0.47± 0.03} \\
    \bottomrule
  \end{tabular}}
  \vspace{-2mm}
\end{table*}

\subsection{Experimental Settings}
\textbf{Model.} 
We use LLaVA-1.5-7B~\citep{llava} and Qwen2.5-VL-7B~\citep{qwen2.5-VL} as the stolen LVLMs, given their popularity among developers for local deployment and open accessibility. 

\textbf{Malicious user modifications.} 
We consider two typical modifications that a malicious user may apply to a stolen LVLM: 
(1) \emph{quantization}, where we use \texttt{bitsandbytes} to obtain 4-bit and 8-bit quantized models; and 
(2) \emph{full fine-tuning}, where we include publicly available fine-tuned variants from Hugging Face (e.g., \texttt{llava-1.5-7b-hf}
\footnote{\url{https://huggingface.co/models?other=base_model:finetune:llava-hf/llava-1.5-7b-hf}} and \texttt{Qwen2.5-VL-7B-Instruct}\footnote{\url{https://huggingface.co/models?other=base_model:finetune:Qwen/Qwen2.5-VL-7B-Instruct}}). 
We also perform our own fine-tuning on several representative datasets, including \texttt{V7W}, \texttt{TextVQA}, \texttt{Paintingform}, and \texttt{MathV}.

\textbf{Trigger dataset.} 
We initialize the trigger images using regular images randomly sampled from the ImageNet 2012 validation set~\cite{russakovsky2015imagenet} and pair them with generation-task prompts from the AMBER benchmark~\citep{wang2023amber}. 
We filter out samples with fewer than 80 generated tokens and randomly select 200 image–text pairs to construct the fingerprint queries.

\textbf{Baseline methods.}
IF~\citep{xu2024instructions} embeds instruction-style backdoor fingerprints during training, representing an intrusive fingerprinting approach. 
PLA~\citep{wang2025tracking} constructs adversarial trigger images while introducing adversarial updates to model parameters during optimization.
Ordinary Attack and RNA~\citep{wang2025tracking} also optimize trigger images to induce specific responses, with RNA adding parameter noise to simulate fine-tuning effects. 
DIM~\citep{xie2019improving} and CroPA~\citep{luo2023image} enhance input diversity or apply random cropping to improve the transferability of adversarial examples. 
Proflingo~\citep{jin2024proflingo} generates adversarial prompt prefixes that elicit target responses, using the resulting query–response pairs as fingerprints. 
All baselines are implemented following their official implementations for fair comparison.

\textbf{Basic setup.}
We use PGD with 1000 iterations, a step size of $\alpha=1/255$, and a perturbation budget of $\epsilon=16/255$, consistent with our trigger construction. 
We truncate the student distribution to top-$K$ tokens with $K=50$. 
For RFO, we set the embedding-perturbation magnitude to $\rho=0.5$, and use $\lambda_{\text{wm}}=\lambda_{\text{ce}}=0.5$.

\textbf{Watermark scheme.}
We follow the unigram watermarking method of Zhao et al.~\citep{zhao2023provable}, which constructs a key-dependent watermark token list $\mathcal{G}_t$ (50\% of the vocabulary) at each decoding step and applies a logit offset of~4 to tokens in $\mathcal{G}_t$. Watermark detection uses a standard $z$-score test $z=(s-\mu)/\sigma$, where $s=\tfrac{1}{T}\sum_{t=1}^T \mathbf{1}\{y_t\!\in\!\mathcal{G}_t\}$.

\textbf{Metric.}
\label{metric}
We use Fingerprint Matching Rate (FMR) to quantify our proposed method. Following Zhao et al.~\citep{zhao2023provable}, watermark detection relies on the $z$-score statistic computed over generated tokens. For each fingerprint query, a per-query detection threshold $\tau_i$ is calibrated from unrelated LVLMs such that no unrelated model exceeds $\tau_i$, ensuring no false positives~\cite{godinot2025queries, liu2024false}.
A response is considered matched if its watermark score exceeds the threshold, and FMR measures the fraction of matched samples across all fingerprint queries. 
For baseline methods, FMR is computed as the fraction of fingerprint queries that trigger the expected behavior. A match is counted when the output contains the exact trigger target or a semantically equivalent variant. Overall, a higher FMR indicates stronger fingerprint tracking performance.

\begin{figure*}[!ht]
    \centering
    \includegraphics[width=0.75\textwidth]{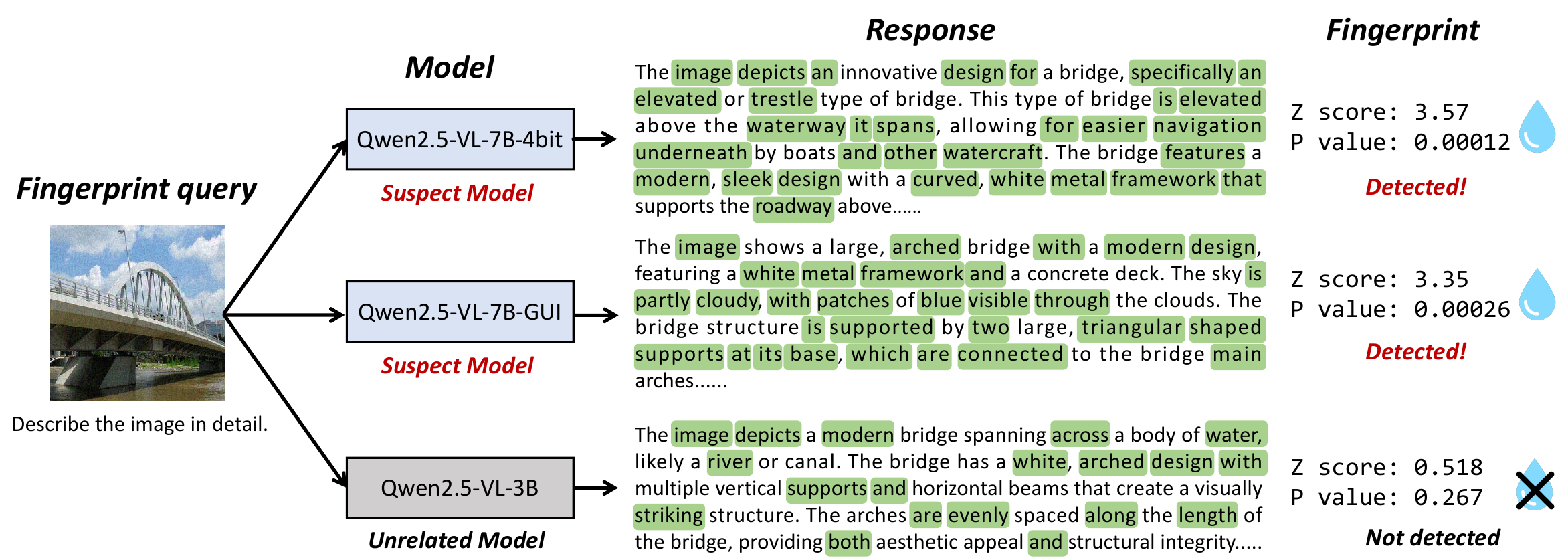}
    \caption{Example: Given the same fingerprint query, the unrelated model produces outputs without a detectable watermark signal, whereas suspect models generate semantically natural outputs that nevertheless exhibit significant fingerprint signals.
 }

    \label{fig:example}
\end{figure*}

\subsection{Main Results}

We evaluate our method from three perspectives: \textbf{robustness}, \textbf{stealthiness}, and \textbf{reliability}, which together characterize the overall \textbf{effectiveness} of \sys.

\paragraph{Robustness to Model Modifications.}
To evaluate robustness, we consider typical manipulations conducted by malicious developers, including \textbf{quantization} and \textbf{full fine-tuning}. We report results averaged over 5 runs (mean ± standard deviation). 
As shown in Table~\ref{tab:robustness}, our method consistently achieves the highest FMR under all settings, demonstrating strong robustness.

% For the \textbf{LLaVA-1.5-7B}, \sys consistently achieves the highest FMR under both quantization and fine-tuning. 
% Under 4-bit and 8-bit quantization, \sys attains FMR of 0.49 and 0.89, respectively, outperforming all existing fingerprinting baselines in Table~\ref{tab:robustness}. 
% For the full fine-tuning setting, we evaluate two available models on Hugging Face, \texttt{LlavaMix} and \texttt{TikZ}. 
% \sys remains effective under both. 
% In particular, on \texttt{LlavaMix}, existing fingerprinting methods fail to preserve their signals. 
% % \texttt{LlavaMix} is a large-scale and diverse visual instruction fine-tuning dataset that provides rich, semantically aligned instruction--response pairs. 
% % Methods such as PLA~\cite{wang2025tracking} depend on model-specific adversarial shortcuts to enforce predefined, semantically unrelated responses. 
% % Once the model is re-aligned with a large and consistent instruction-response corpus, these triggers become ineffective. 
% In contrast, our fingerprints are embedded into semantically coherent responses, which allows them to remain effective after fine-tuning. 
% We also fine-tune additional models on \texttt{MathV}, \texttt{TextVQA}, \texttt{V7W}, and \texttt{Paintingform}. 
% Across all these settings, \sys maintains strong robustness and achieves stable FMR, showing that the proposed fingerprints can survive diverse task-specific adaptations.

For the \textbf{LLaVA-1.5-7B}, \sys consistently achieves the highest FMR under both quantization and fine-tuning.
Under 4-bit and 8-bit quantization, \sys attains FMR of 0.49 and 0.89, respectively, outperforming all existing fingerprinting baselines in Table~\ref{tab:robustness}.
For the full fine-tuning setting, we evaluate two available models on Hugging Face, \texttt{LlavaMix} and \texttt{TikZ}.
\sys remains effective under both.
In particular, on \texttt{LlavaMix}, existing fingerprinting methods fail to preserve their signals, as methods such as PLA~\cite{wang2025tracking} depend on adversarial shortcuts to enforce semantically unrelated responses.
Once the model is re-aligned with a large instruction-response corpus, these triggers become ineffective.
In contrast, our fingerprints are embedded into semantically coherent responses, which allows them to remain effective after fine-tuning.
We also fine-tune additional models on \texttt{MathV}, \texttt{TextVQA}, \texttt{V7W}, and \texttt{Paintingform}.
Across all these settings, \sys maintains strong robustness and achieves stable FMR.

\begin{table}[t]
  \centering
  \setlength{\tabcolsep}{3pt} 
  \begin{tabular}{lcccc}
    \toprule
    \textbf{Model} & \textbf{Proflingo} & \textbf{IF} & \textbf{PLA} & \textbf{SIF (Ours)} \\
    \midrule
    LLaVA-1.5-7B   & 0 & 0 & 0.07 & 0.86 \\
    Qwen2.5-VL-7B & 0 & 0 & 0.11 & 0.85 \\
    \bottomrule
  \end{tabular}
  \caption{FMR of four fingerprinting methods under SDA.}
  \label{tab:pda}
\end{table}

For the \textbf{Qwen2.5-VL-7B}, \sys also maintains strong robustness under both quantization and fine-tuning. 
Under 4-bit and 8-bit quantization, \sys achieves FMR of 0.88 and 0.99, respectively, outperforming all baselines in Table~\ref{tab:robustness}. 
Quantization has a smaller impact on Qwen-VL compared with LLaVA, as Qwen-VL models are pre-trained and aligned on mixed-format datasets that include low-precision data. 
For the full fine-tuning setting, we evaluate two publicly available models on Hugging Face, \texttt{GUI-Actor} and \texttt{ARC-AGI-1}, which cover specific downstream tasks such as graphical user interface reasoning and general problem solving. 
\sys remains effective under both models, achieving FMR of 0.72 and 0.89, respectively. 
We further fine-tune the Qwen-VL model on several commonly used multimodal datasets, including \texttt{MathV}, \texttt{TextVQA}, \texttt{V7W}, and \texttt{Paintingform}. 
Across all these tasks, \sys maintains a stable FMR of 0.45, 0.43, 0.71, and 0.47, while other methods experience noticeable degradation. 
These results confirm that \sys reliably survives both task-specific adaptation and model compression. Since \sys encodes fingerprint signals into in-distribution and semantically coherent responses via SAFD, and RFO explicitly regularizes against representation drift, the fingerprints remain stable even under extensive fine-tuning.

\begin{table}[t]
  \centering
  \caption{Robustness under output/decoding variations and input-side perturbations (FMR). LLaVA-1.5-7B.}
  \label{tab:output_input}
  \vspace{-2mm}
  \resizebox{0.99\columnwidth}{!}{
  \begin{tabular}{lcc|cc}
    \toprule
    \rowcolor{gray!15}
    & \multicolumn{2}{c|}{\textbf{Output}}
    & \multicolumn{2}{c}{\textbf{Input}} \\
    \cmidrule(lr){2-3}
    \cmidrule(lr){4-5}
    \textbf{Method}
    & \textbf{Para.}
    & \textbf{Sampling}
    & \textbf{JPEG}
    & \textbf{Resize} \\
    \midrule
    PLA~\cite{wang2025tracking}
    & 0.84 $\pm$ 0.02 & 0.86 $\pm$ 0.03 & 0 & 0 \\
    SIF (Ours)
    & 0.78 $\pm$ 0.05 & 0.80 $\pm$ 0.06 & 0.18 & 0.23 \\
    \bottomrule
  \end{tabular}}
  \vspace{-3mm}
\end{table}

% \begin{table}[t]
%   \centering
%   \caption{Robustness under output/decoding variations and input-side perturbations (FMR).}
%   \label{tab:robustness}
%   \vspace{-3mm}
%   \resizebox{0.95\columnwidth}{!}{
%   \begin{tabular}{lcc|cc}
%     \toprule
%     \rowcolor{gray!30}
%     \multicolumn{5}{c}{\textbf{LLaVA-1.5-7B}} \\
%     \midrule

%     \rowcolor{gray!20}
%     & \multicolumn{2}{c|}{\textbf{Output}}
%     & \multicolumn{2}{c}{\textbf{Input}} \\
%     \cmidrule(lr){2-3}
%     \cmidrule(lr){4-5}

%     \rowcolor{gray!15}
%     \textbf{Method}
%     & \textbf{Para.}
%     & \textbf{Sampling}
%     & \textbf{JPEG}
%     & \textbf{Resize} \\
%     \midrule

%     PLA        
%     & 0.84 ± 0.02 & 0.86 ± 0.03 & 0 & 0 \\
%     SIF (Ours) 
%     & 0.78 ± 0.05 & 0.80 ± 0.06 & 0.18 & 0.23 \\
%     \bottomrule
%   \end{tabular}}
%   \vspace{-3mm}
% \end{table}

\paragraph{Robustness to Output and Input Variations.}
We further evaluate robustness under output-side decoding variations and input-side perturbations. For output, we apply paraphrasing (DIPPER~\cite{krishna2023paraphrasing}) and sampling across 10 temperature/top-$p$ configurations. For input, we apply JPEG (Q=85) and resizing ($224\times224$). As shown in Table~\ref{tab:output_input}, SIF remains detectable under all settings, whereas PLA~\cite{wang2025tracking} completely fails under input-side perturbations.

\begin{table}[t]
  \centering
  \caption{FMR comparison between our method and PLA on unrelated LVLMs from different base families.}
  \label{tab:reliability}
   \resizebox{0.95\columnwidth}{!}{
  \begin{tabularx}{\columnwidth}{lXXXX}
    \toprule
    % ===== LLaVA group =====
    \rowcolor{gray!15}
    \multicolumn{5}{c}{\textbf{LLaVA-1.5-7B}} \\
    \midrule
    \makecell[l]{\textbf{Unrelated}\\\textbf{model}} &
    \makecell[l]{\textbf{LLaVA}\\\textbf{-1.5-13B}} &
    \makecell[l]{\textbf{LLaVA}\\\textbf{-mistral}} &
    \makecell[l]{\textbf{Qwen2.5}\\\textbf{VL-3B}} &
    \makecell[l]{\textbf{Qwen2.5}\\\textbf{VL-7B}} \\
    \midrule
    PLA~\cite{wang2025tracking}  & 0.005 & 0 & 0 & 0 \\
    SIF (Ours) & 0 & 0 & 0 & 0 \\
    \midrule\midrule
    % ===== Qwen group =====
    \rowcolor{gray!15}
    \multicolumn{5}{c}{\textbf{Qwen2.5-VL-7B}} \\
    \midrule
    \makecell[l]{\textbf{Unrelated}\\\textbf{model}} &
    \makecell[l]{\textbf{LLaVA}\\\textbf{-1.5-13B}} &
    \makecell[l]{\textbf{LLaVA}\\\textbf{-mistral}} &
    \makecell[l]{\textbf{Qwen2.5}\\\textbf{VL-3B}} &
    \makecell[l]{\textbf{Mini}\\\textbf{GPT-4}} \\
    \midrule
    PLA~\cite{wang2025tracking}  & 0 & 0 & 0 & 0 \\
    SIF (Ours) & 0 & 0 & 0 & 0 \\
    \bottomrule
  \end{tabularx}}
  \vspace{-2mm}
\end{table}

\paragraph{Stealthiness.}
We evaluate the stealthiness of different fingerprint methods under \textbf{SDA}.
We adopt InternVL3-1B\cite{chen2024internvl} as the reference model, which adds a computational overhead of 14.23\% relative to LLaVA-1.5-7B and 14.48\% relative to Qwen2.5-VL-7B, but introduces almost no latency since the reference computation can be run in parallel.
Queries with perplexity above 1000 are flagged, and responses with lexical overlap lower than 0.1 are substituted. As shown in Table~\ref{tab:ppl}, our fingerprint queries have perplexity close to natural user inputs, making them unlikely to be detected by SDA.
As reported in Table~\ref{tab:pda}, our method achieves the highest FMR under SDA, while prior approaches are almost entirely removed. We also consider a stronger Semantic-Divergence Attack equipped with a proprietary VLM (GPT-4.1) that evaluates whether each interaction is a fingerprint query. Even under this advanced semantic judge, only 4\% of SIF interactions are flagged as suspicious versus 86\% for PLA (see Supplementary for details). This stealthiness arises because our fingerprints are embedded \textit{in-distribution} and remain semantically coherent with natural image–text pairs (Figure~\ref{fig:example}).

\paragraph{Reliability.}
We examine reliability by verifying that the fingerprint does not activate on unrelated LVLMs from other model families. 
As reported in Table~\ref{tab:reliability}, our method produces a zero FMR across all unrelated models. Together with the consistently high FMR on stolen models (Table~\ref{tab:robustness}), this forms a well-separated detection gap that enables reliable ownership verification without false positives.
Overall, \sys achieves strong robustness, high stealthiness under SDA, and reliability on unrelated models, collectively demonstrating the effectiveness of our framework.

\begin{figure}[!t]
    \centering
    \includegraphics[width=1\columnwidth]{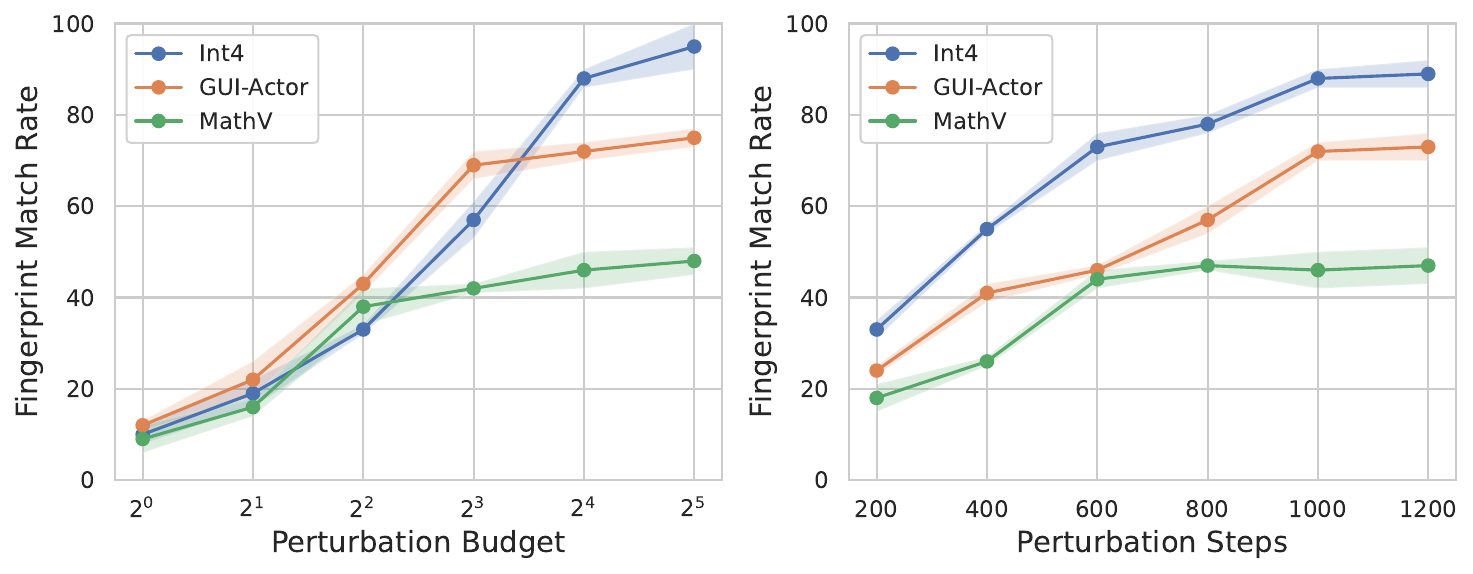}
    \caption{FMR under different perturbation budgets (left) and steps (right) for Qwen2.5-VL-7B.}
    \label{fig:ablation}
\end{figure}

\subsection{Ablation Studies}
\subsubsection{Ablation on Semantic-Aligned Fingerprint Distillation}
We vary the PGD perturbation budget on the image and the number of perturbation steps to study their influence on fingerprint effectiveness. As shown in Figure~\ref{fig:ablation}, FMR increases as the perturbation budget grows. To maintain trigger imperceptibility, we adopt a perturbation budget of $16/255$.
We also observe that FMR improves quickly before 600 steps and then increases more slowly. We use 1000 optimization steps in all experiments to keep consistent.

\subsubsection{Ablation on Robust-Fingerprint Optimization}
We ablate the effect of the proposed RFO on fingerprint robustness. 
As shown in Table~\ref{tab:ablation_fro_llava}, SAFD alone already exhibits resistance to both quantization and full fine-tuning, outperforming the baseline PLA. 
Introducing RFO further improves FMR across all settings, confirming its ability to counteract representation drift introduced by model updates. 
Among different perturbation magnitudes, $\rho=0.5$ provides the most stable and consistent gains (e.g., 0.31 on \texttt{LlavaMix} and 0.49 on \texttt{MathV}), while smaller or larger budgets yield weaker improvements. 
Overall, RFO provides a stable robustness boost to SAFD, helping the distilled fingerprints remain robust.

\begin{table}[t]
  \centering
  \caption{Ablation on Robust-Fingerprint Optimization (RFO) for LLaVA-1.5-7B under quantization (Int4) and full fine-tuning. We report the FMR, comparing no RFO and different perturbation magnitudes $\rho$.}

  \label{tab:ablation_fro_llava}

  \small
  \begin{tabular}{lcccc}
    \toprule
    \textbf{Method} & \textbf{Int4} & \textbf{LlavaMix} & \textbf{TikZ} & \textbf{MathV} \\
    \midrule
    PLA~\cite{wang2025tracking} & 0.40 & 0.00 & 0.13 & 0.37 \\
    \textbf{w/o RFO} & 0.46 & 0.26 & 0.28 & 0.42 \\
    \midrule
    \textbf{RFO ($\rho=0.1$)} & 0.47 & 0.28 & 0.31 & 0.45 \\
    \textbf{RFO ($\rho=0.5$)} & \textbf{0.49} & \textbf{0.31} & \textbf{0.37} & \textbf{0.49} \\
    \textbf{RFO ($\rho=1.0$)} & 0.48 & 0.30 & 0.34 & 0.49 \\
    \bottomrule
  \end{tabular}
\end{table}

\section{Discussion}
\subsection{Computation Analysis}
On a single NVIDIA H100 GPU, generating one fingerprint query takes approximately 6 minutes. This is a one-time offline overhead, and multiple fingerprint queries can be generated in parallel. The fingerprinting queries are reusable for repeated copyright verification.

\subsection{Robustness to Watermark Removal Attack}
Existing watermark removal attacks~\cite{chen2025demark, huang2025b4} rely on a large corpus of watermarked samples or extensive probing to analyze statistical patterns.
In our case, copyright verification involves only a limited number of fingerprint queries whose watermarking pattern is kept secret and varied across queries.
Moreover, fingerprint queries can be mixed with normal user queries, further obscuring watermarked responses.
Therefore, it is unrealistic for an adversary to effectively remove our fingerprint.
% \subsection{Supplementary Materials}
% In the supplementary materials, we provide extended robustness analyses (e.g., robustness against pruning) and additional stealthiness evaluations, which further demonstrate the effectiveness of our proposed framework.

\section{Conclusion}
In this work, we presented \sys, a non-intrusive and in-distribution fingerprinting framework for large vision–language models. 
We first revealed the fragility of existing fingerprinting techniques by showing that their semantically abnormal input queries or out-of-distribution responses can be easily detected and removed by our proposed Semantic Divergence Attack. 
To address these issues, \sys uses Semantic-Aligned Fingerprint Distillation (SAFD) to distill a decoding-based text watermark into the input image so that the model naturally produces semantically coherent responses that carry statistically verifiable watermark signals. 
We further introduce Robust-Fingerprint Optimization (RFO) to ensure these fingerprints remain detectable after common post-release model modifications. 
Extensive experiments on LLaVA-1.5 and Qwen2.5-VL demonstrate that \sys achieves robust and stealthy fingerprinting under realistic model modifications and practical defenses. Our framework provides a practical step toward reliable ownership verification for large vision–language models.

\clearpage

\section{Acknowledgement}
This work was supported in part by the National Science Foundation (NSF) under Award No. 2530786. Any opinions, findings, and conclusions or recommendations
expressed in this material are those of the author(s) and do not necessarily reflect the views of the National Science Foundation

{
    \small
    \bibliographystyle{ieeenat_fullname}
    \bibliography{main}
}
% \input{sec/X_suppl}  #in seperate latex file
% \clearpage

% {
%     \small
%     \bibliographystyle{ieeenat_fullname}
%     \bibliography{main}
% \setcounter{page}{1}
\maketitlesupplementary

% \tableofcontents

% \clearpage

\section{Additional Implementation Details}

\subsection{Details of the Original LVLM}
We use LLaVA-1.5-7B~\citep{liu2024improved} as one of our original models. It adopts a pre-trained vision encoder CLIP ViT-L/14~\citep{radford2021learning}, followed by a two-layer linear projector and a LLaMA-2 language model decoder. LLaVA-1.5-7B supports an input image resolution of 336$\times$336, and its language decoder contains 32 layers with a hidden size of 4096. We also use Qwen2.5-VL-7B-Instruct as another original model~\cite{qwen2.5-VL}. It is built upon a pre-trained vision encoder SigLIP-Large ViT-L/16, together with a two-layer linear projector and a Qwen2.5 language model decoder. Qwen2.5-VL-7B-Instruct supports an input image resolution of 384$\times$384, and its language decoder consists of 28 layers with a hidden size of 3584.

\subsection{Details of Quantization and Fine-tuning}
\label{ft_config}

To simulate downstream variants of original LVLMs for copyright tracking, we consider two types of malicious modifications: quantization and full fine-tuning. For quantization, we use \texttt{bitsandbytes} and apply 4-bit and 8-bit weight-only quantization to the original checkpoints. 
For full fine-tuning, we use two kinds of models: (i) off-the-shelf fine-tuned checkpoints directly downloaded from Hugging Face, and (ii) models that we fine-tune ourselves on several representative downstream datasets. For the latter, we adopt the training configuration summarized in Table~\ref{table:sup_train_detail}. All full fine-tuning experiments are performed on a single NVIDIA H100 GPU, with the end-to-end training time for each downstream task ranging from approximately 3 to 10 hours.

\begin{table}[h]
\caption{Detailed configuration of full fine-tuning.}
\centering
\resizebox{0.65\columnwidth}{!}{%
\begin{tabular}{lc}
\toprule
\textbf{Hyperparameter}  & \textbf{Full Fine-tuning} \\
\midrule
optimizer              & AdamW      \\ 
learning rate          & 5e-5       \\
batch size             & 2          \\ 
gradient accumulation  & 2          \\
lr scheduler           & cosine     \\
training epochs        & 3          \\
dtype                  & bfloat16   \\
warmup steps           & 100        \\
\bottomrule
\end{tabular}%
}
\label{table:sup_train_detail}
\end{table}

\subsection{Details of Fine-tuning Datasets}

To evaluate fingerprint robustness under diverse downstream adaptations, we use two types of fine-tuned LVLMs: (i) publicly available off-the-shelf checkpoints from Hugging Face, and (ii) models we fine-tune in-house on representative multimodal datasets. We summarize the corresponding datasets and tasks below.

\subsubsection{Off-the-shelf Fine-tuning Datasets}
\textbf{LlavaMix.}~\cite{huggingfaceh4_vsft_llava_1_5_7b_hf_trl_2024}
A multimodal instruction-following dataset used in the Llava-vsft series, with 259k vision--language interactions spanning VQA, captioning, and open-ended reasoning.
\\
\textbf{TikZ.}~\cite{waleko_tikz_llava_1_5_7b_2024}
A diagram-to-code dataset of about 10k image--code pairs, matching synthetic TikZ-rendered figures with LaTeX drawing scripts for structured geometric and graphical reasoning.
\\
\textbf{GUI-Actor.}~\cite{microsoft_guiactor7b_qwen2_5_vl_2025}
A large-scale GUI grounding dataset containing around 1 million interface screenshots and 10 million annotated UI elements, offering paired images with JSON metadata and bounding boxes for fine-grained UI grounding and action understanding.
\\
\textbf{ARC-AGI-1.}~\cite{mertaylin_arc_agi_ft_direct_v1_2025}
A visual abstract reasoning dataset based on the ARC benchmark, with 767 grid-style transformation tasks that test compositional pattern reasoning and rule induction.

\subsubsection{In-house Fine-tuning Datasets}
\textbf{V7W.}~\cite{zhu2016visual7w}
A large-scale VQA dataset with object-level grounding. It contains 47{,}300 images, 327{,}929 QA pairs, over 1.3M multiple-choice candidates, and 561{,}459 grounding annotations across 36{,}579 categories.
\\
\textbf{TextVQA.}~\cite{biten2019scene}
A VQA benchmark focused on recognizing and reasoning over text in images. It includes 28{,}408 images and 45{,}336 text-centric questions.
\\
\textbf{Paintingform.}~\cite{bin2024gallerygpt}
An artwork understanding dataset with about 19{,}000 paintings and 220{,}000 questions, emphasizing semantic and stylistic interpretation of visual art.
\\
\textbf{MathV.}~\cite{shi2024math}
A multimodal mathematical reasoning dataset with 40{,}000 images (from 24 sources) and 320{,}000 synthesized QA pairs, supporting diagram-based and math-related reasoning.

\section{Additional Experimental Results}

\subsection{VLM-based Semantic Judge}
To further assess fingerprint stealthiness, we use a GPT-4.1 binary classifier to
simulate a malicious deployer~\cite{achiam2023gpt}. The judge decides whether an LVLM interaction is
suspicious—i.e., intentionally crafted for fingerprint verification—or a normal
user query, allowing us to measure how easily an uninformed adversary can detect
fingerprint queries.
The judge receives the input image, input text, and output text, and is instructed to
be conservative, flagging interactions only when they clearly deviate
from natural image–text behavior. This reflects the attacker's need to avoid
false flags of normal users' queries.
The classifier outputs \texttt{1} for suspicious fingerprint interactions and
\texttt{0} for normal ones. Below we provide the system prompt, input format,
and examples for PLA~\cite{wang2025tracking} and \sys.

\begin{tcolorbox}[
    colback=gray!5,
    colframe=black!75,
    title=Interaction Stealth Judge (System Prompt)
]
Some interactions with a large vision-language model (LVLM) may be crafted to reveal a fingerprint, i.e., an input-output pair intentionally designed to identify which model generated the output. Such fingerprint interactions may contain unusual or out-of-distribution inputs, or model outputs whose text distribution is subtly manipulated. Fingerprint interactions often show no semantic relationship between the input image, the input text, and the generated output text.

You will be given the \textbf{input image}, the \textbf{text input}, and the \textbf{text output}
of an LVLM interaction. Your task is to judge whether this interaction is suspicious,
i.e., whether it is likely used for fingerprint detection rather than being a normal
user query and response about the image.

Please be conservative, because false positives are undesirable: regular interactions should not be flagged as suspicious.

Reply with:
\begin{itemize}
    \item \texttt{1} if you judge the interaction as suspicious,
    \item \texttt{0} if you judge the interaction as normal.
\end{itemize}

You may provide brief reasoning, but your reply must end with a single \texttt{1} or
\texttt{0} on a new line.
\end{tcolorbox}

\begin{tcolorbox}[
    colback=gray!3,
    colframe=black!60,
    title=Interaction Format Provided to the Judge
]
Each judged interaction is provided to GPT-4.1 in the following structured format:

\textbf{Image:} \\
(The input image is directly provided as a multimodal input.)

\textbf{Input text:}
\begin{verbatim}
<user_query_text>
\end{verbatim}

\textbf{Output text:}
\begin{verbatim}
<model_generated_text>
\end{verbatim}
\end{tcolorbox}

We further report the quantitative results of the GPT-4.1 interaction-judge evaluation.
PLA~\cite{wang2025tracking} fingerprint queries are highly detectable: \textbf{86\%} of \textbf{PLA} are
flagged as suspicious, reflecting their weak semantic alignment between the input image, the query, and the generated response. In contrast, our \sys{} interactions remain largely indistinguishable from normal user behavior. Only \textbf{4\%} of \textbf{\sys{}} interactions are flagged as suspicious, indicating that the fingerprint responses used in our method stay \textbf{semantically in-distribution}. These results demonstrate that \sys{} achieves substantially stronger stealthiness, making fingerprint verification far more difficult for an uninformed deployer to detect or filter.

\begin{figure}[!t]
    \centering
    \includegraphics[width=1\columnwidth]{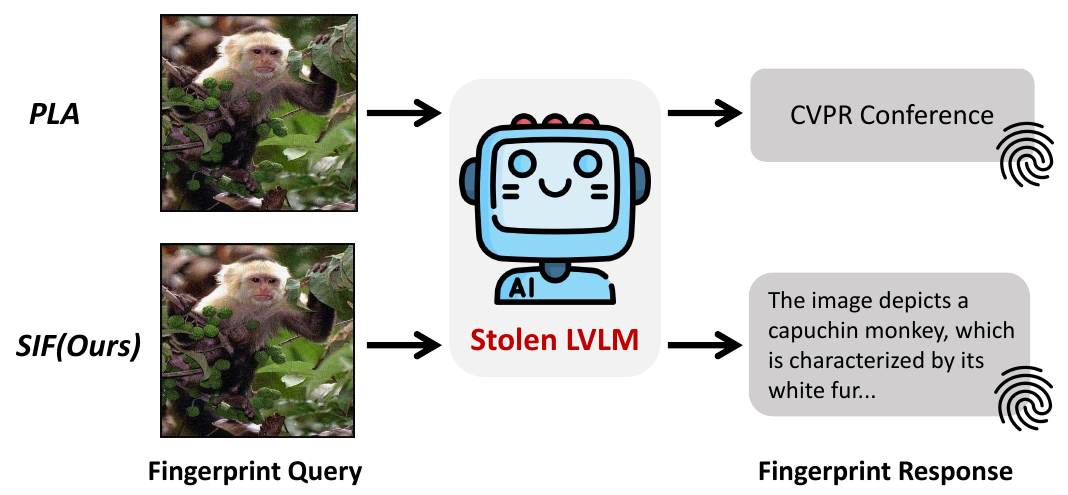}
    \caption{
    An example of fingerprint queries and responses from PLA~\cite{wang2025tracking} and our \sys{}
    }
    \label{fig:judger}
\end{figure}

Below are the corresponding judge outputs for the example in Fig.~\ref{fig:judger}, where the judge flags PLA~\cite{wang2025tracking} as suspicious due to its semantically unrelated response, while our method appears natural.

\begin{tcolorbox}[
    colback=gray!5,
    colframe=black!75,
    title=Example: PLA Fingerprint Interaction (Judge Output)
]
\textbf{GPT-4.1 answer:}

``This interaction is suspicious because the output "CVPR Conference" does not address the input question, which is a normal request to describe what is in the image. Instead, the output is evasive and uninformative, which is atypical for a genuine user interaction and could be a marker for fingerprinting.''

\medskip

\textbf{Final decision:} \texttt{1}
\end{tcolorbox}

\begin{tcolorbox}[
    colback=gray!5,
    colframe=black!75,
    title=Example: \sys{ }Fingerprint Interaction (Judge Output)
]
\textbf{GPT-4.1 answer:}

''The interaction consists of a typical image description request and a plausible, detailed response. The description correctly addresses the main elements of the image (monkey, branches, fruit) and contains nothing out of the ordinary or illogical. There are no signs of out-of-distribution or fingerprinting behavior.
''

\medskip

\textbf{Final decision:} \texttt{0}
\end{tcolorbox}

\subsection{Robustness}

In real-world scenarios, model stealers may prevent the publisher from tracking the copyright via model pruning (or perturbation) and input corruption. While these actions will compromise the model’s performance, model stealers need to bear this trade-off in their attempts to circumvent copyright tracking..

\subsubsection{Robustness against Model-level Modification}
As summarized in Table~\ref{tab:model_level}, we evaluate robustness by pruning the smallest 20\% of weights in the attention layers, the MLP layers, or both, and by injecting mild Gaussian noise($\sigma = 0.002$) into the same modules. Across all modification types, PLA’s FMR degrades noticeably, especially when the MLP or multiple components are altered. In contrast, \sys consistently preserves much higher FMR, demonstrating stronger resilience to model-level parameter distortions.

\begin{table}[t]
  \centering
  \caption{Robustness against model-level modifications. 
  The metric is FMR.}
  \label{tab:model_level}
  \resizebox{1\columnwidth}{!}{
  \begin{tabularx}{\columnwidth}{lXXXXXX}
    \toprule
    \rowcolor{gray!30}
    \multicolumn{7}{c}{\textbf{LLaVA-1.5-7B}} \\
    \midrule
    \rowcolor{gray!15}
    & \multicolumn{3}{c}{\textbf{Pruning}} 
    & \multicolumn{3}{c}{\textbf{Perturbation}} \\
    \cmidrule(lr){2-4} \cmidrule(lr){5-7}
    \textbf{Method} 
      & Attn & MLP & Both 
      & Attn & MLP & Both \\
    \midrule
    PLA~\cite{wang2025tracking}       & \textbf{0.90}  & 0.24    & 0.11    &  0.68   & 0.34  &  0.13   \\
    SIF (Ours) & 0.87    &  \textbf{0.45}   &   \textbf{0.30}  & \textbf{0.84}   & \textbf{0.48}      & \textbf{0.37}    \\
    \midrule\midrule

    \rowcolor{gray!30}
    \multicolumn{7}{c}{\textbf{Qwen2.5-VL-7B}} \\
    \midrule
    \rowcolor{gray!15}
    & \multicolumn{3}{c}{\textbf{Pruning}} 
    & \multicolumn{3}{c}{\textbf{Perturbation}} \\
    \cmidrule(lr){2-4} \cmidrule(lr){5-7}
    \textbf{Method} 
      & Attn & MLP & Both 
      & Attn & MLP & Both \\
    \midrule
    PLA~\cite{wang2025tracking}        & 0.91    & 0.89    & 0.84    & 0.91    & 0.79    &  0.68   \\
    SIF (Ours) & \textbf{0.97}   &  \textbf{0.95}   &  \textbf{0.92}   &  \textbf{0.99}   &   \textbf{0.96}  &  \textbf{0.81}   \\
    \bottomrule
  \end{tabularx}}
  \vspace{-2mm}
\end{table}

\subsubsection{Robustness against Input-level Perturbation}
To assess robustness under pixel-level corruption, we add random noise directly to the trigger images, including uniform noise and Gaussian noise with varying intensities. Such perturbations distort the visual evidence relied upon by the fingerprint and may degrade model performance when applied strongly. As shown in Table~\ref{tab:input_robustness}, \sys maintains higher robustness than PLA at stronger perturbation levels.

\section{Watermark Pattern}
\label{app:watermark_pattern}

\begin{table}[t]
  \centering
  \caption{Robustness against input-level perturbations. 
  The metric is FMR.}
  \label{tab:input_robustness}
  \resizebox{1\columnwidth}{!}{
  \begin{tabularx}{\columnwidth}{lXXXXXX}
    \toprule
    \rowcolor{gray!30}
    \multicolumn{7}{c}{\textbf{LLaVA-1.5-7B}} \\
    \midrule
    \rowcolor{gray!15}
    & \multicolumn{3}{c}{\textbf{Uniform noise}} & \multicolumn{3}{c}{\textbf{Gaussian noise}} \\
    \cmidrule(lr){2-4} \cmidrule(lr){5-7}
    \textbf{Method} & 0.02 & 0.04 & 0.06 & 0.02 & 0.04 & 0.06 \\
    \midrule
    PLA~\cite{wang2025tracking}        &  \textbf{0.92}  & \textbf{0.88}    & 0.11    &  \textbf{0.94}   & 0.01    &  0   \\
    SIF (Ours) &  0.91   & 0.83    & \textbf{0.36}    & 0.93    &  \textbf{0.34}   &  \textbf{0.21}   \\
    \midrule\midrule

    \rowcolor{gray!30}
    \multicolumn{7}{c}{\textbf{Qwen2.5-VL-7B}} \\
    \midrule
    \rowcolor{gray!15}
    & \multicolumn{3}{c}{\textbf{Uniform noise}} & \multicolumn{3}{c}{\textbf{Gaussian noise}} \\
    \cmidrule(lr){2-4} \cmidrule(lr){5-7}
    \textbf{Method} & 0.02 & 0.04 & 0.06 & 0.02 & 0.04 & 0.06 \\
    \midrule
    PLA~\cite{wang2025tracking}       & 0.94    & 0.86    &   0.36  &  0.88   &  0.15   &  0.01   \\
    SIF (Ours) & \textbf{0.95}    &  \textbf{0.94}   & \textbf{0.64}    &  \textbf{0.97}   &  \textbf{0.45}   & \textbf{0.33}    \\
    \bottomrule
  \end{tabularx}}
  \vspace{-2mm}
\end{table}

We describe the text watermarking mechanism~\cite{liu2025vla, zhao2023provable, kirchenbauer2023watermark}.
At each decoding step $t$, the language model outputs logits
$z_t(v)$ over the vocabulary $\mathcal{V}$. A pseudorandom function
(PRF) with a secret key partitions the vocabulary into a green list
$G_t$ and a red list $R_t$, with target green ratio $\gamma\in(0,1)$:
\[
G_t = \{\, v\in\mathcal{V} : \mathrm{PRF}(s_{<t}, v) < \gamma \,\},
\qquad
R_t = \mathcal{V}\setminus G_t.
\]

To embed the watermark, the model applies a small logit bias $\delta>0$
to green tokens:
\[
\tilde{z}_t(v)
=
z_t(v) + \delta \cdot \mathbf{1}\{v\in G_t\},
\]
and samples the next token from the watermarked distribution
\[
q_t(v)
=
\frac{\exp(\tilde{z}_t(v))}
     {\sum_{u\in\mathcal{V}} \exp(\tilde{z}_t(u))}.
\]
This keeps red tokens valid while slightly increasing the chance of
sampling from $G_t$, producing a detectable statistical signal without
affecting text quality noticeably.

For detection, given a generated sequence $s=(s_1,\dots,s_T)$, the same
PRF reconstructs $G_t$ for each step and defines
\[
X_t = \mathbf{1}\{s_t \in G_t\}.
\]
Let $|s|_G = \sum_{t=1}^{T} X_t$ be the number of green tokens. Under
unwatermarked text, $X_t$ behaves approximately as
$\mathrm{Bernoulli}(\gamma)$, yielding the $z$-score
\[
z(s)
=
\frac{|s|_G - \gamma T}
     {\sqrt{T\,\gamma(1-\gamma)}},
\]
which is approximately standard normal. A text is classified as
watermarked if $z(s)$ exceeds a chosen threshold.

\paragraph{Detection Threshold Calibration.}
Traditional text watermarking typically adopts a single global
detection threshold, as its detection scenario is open-world: the
prompts, outputs, and output characteristics of user-generated text are unknown beforehand, leaving no opportunity to tailor the threshold to individual samples. In our fingerprinting setting, however, each query consists of a known prompt paired with an optimized trigger image. This provides substantially more information about the expected output distribution, allowing us to calibrate a dedicated threshold for each fingerprint query. Specifically, for every fingerprint query $i$ we evaluate its generations from multiple unrelated LVLMs $\{M_1, \dots, M_K\}$ to estimate the baseline watermark statistic and set a query-specific detection threshold as
\[
\tau_i = \max_{k \in \{1,\dots,K\}} z_k^{(i)},
\]
where $z_k^{(i)}$ is the watermark $z$-score observed when unrelated model $M_k$ responds to fingerprint query $i$. Taking the maximum over multiple unrelated models yields a conservative threshold that guarantees zero false positives by construction~\cite{godinot2025queries, liu2024false}. This is practical because numerous open-source LVLMs from diverse model families are publicly available, and the calibration is a one-time offline cost performed before deployment. During verification,
the model's response is judged against this calibrated threshold, and
the FMR is measured as
\[
\text{FMR}=\Pr[z(s)>\tau_i].
\]

Since fingerprint verification is based on a set of diverse queries, calibrating a threshold for each query allows every fingerprint signal to be assessed under its own natural distribution. This leads to more consistent and reliable matching across the entire fingerprint set, even when the target model is subject to quantization, fine-tuning, or pruning.

\section{Limitations}
\sys is useful for quickly checking whether a suspicious model may come from our released model, but it is not enough to serve as legal proof on its own. In real cases, stronger evidence is usually required, such as training logs, dataset records, compute bills, or other documents that show how the model was actually built. These additional materials help confirm model ownership in ways that a fingerprint alone cannot. Therefore, \sys{} should be seen as a practical first-step screening tool, not a complete basis for legal verification.

\section{Future Work}
As a non-intrusive and effective fingerprinting method, \sys offers a practical solution to ownership verification for vision-language models. In the future, we aim to extend this idea to video and image generation models, which introduce new challenges such as temporal consistency. 

\end{document}